# An Improved Image Mosaicing Algorithm for Damaged Documents

Er. Waheeda Dhokley
Computer dept. university of Mumbai
M.H Saboo Siddik College of Engineering
Mumbai, India

Shaikh Nazia
Computer dept. university of Mumbai
M.H Saboo Siddik College of Engineering
Mumbai, India

Khan Munifa
Computer dept. university of Mumbai
M.H Saboo Siddik College of Engineering
Mumbai, India

Shaikh Saiqua
Computer dept. university of Mumbai
M.H Saboo Siddik College of Engineering
Mumbai, India

*Abstract*—It is a common phenomenon in day to day life; where in some of the document gets damaged. Out of several reasons, the main reason for documents getting damaged is shredding by hands. Recovery of such documents is essential. Manual recovery of such damaged document is tedious and time consuming task. In this paper, we are describing an algorithm which recovers the original document from such shredded pieces of the same. In order to implement this, we are using a simple technique called Image Mosaicing. In this technique a complete new image is developed using two or more torn fragments. For simplicity of implementation, we are considering only two torn pieces of a document that will be mosaiced together. The successful implementation of this algorithm would lead to recovery of important information which in turn would be beneficial in various fields such as forensic sciences, archival study, etc

*Keywords- Image Mosaicing, Torn fragments*

## I. INTRODUCTION

During the course of investigation in fields such as forensics sciences and archival study, there is loss of valuable information occurring due to torn documents. Documents often get damaged or torn not only because of aging but also due to shredding of documents intentionally or un-intentionally. For small scale problems, any manual random search can be used, whereas in case of large scale problems, due to increased complexity, using this methodology proves to be tedious and time consuming task.

To overcome these large scale problems encountered during documents recovery, technique of image of Image Mosaicing is used. Image mosaicing is a process of reconstructing or stitching single, continuous image from a set of separate or overlap sub images. Image mosaicing is one of the crucial steps for automatically reconstructing damaged documents. The main benefit of image mosaicing is that there is no involvement of physical reconstructions of a document (i.e. using glue, adhesives, etc). Mosaicing is free from destructive analysis. Image mosaicing is a methodology which is easy to implement.

The main focus of our project is to implement recovery of shredded documents with accurate result. For this implementation in our project we are considering two pieces or to fragments of damaged document to be mosaiced together. This is achieved using various steps. The input for our project is two torn fragments, which are scanned. These data is then passed onto various algorithm which results to a continuous mosaiced image.

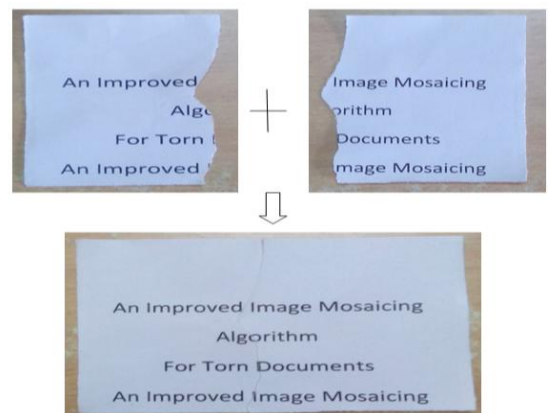

Figure 1: Recovery Of Original Document Using Two Torn pieces

At times, due to certain unavoidable actions, certain information is lost which can not be recovered even after mosaicing. Examples may include part of paper including some letters of a word is destroyed. Recovering such document may result in incomplete word formation. In order to overcome this problem, character recognition is used so that the word can be recognized and the original word can be written.

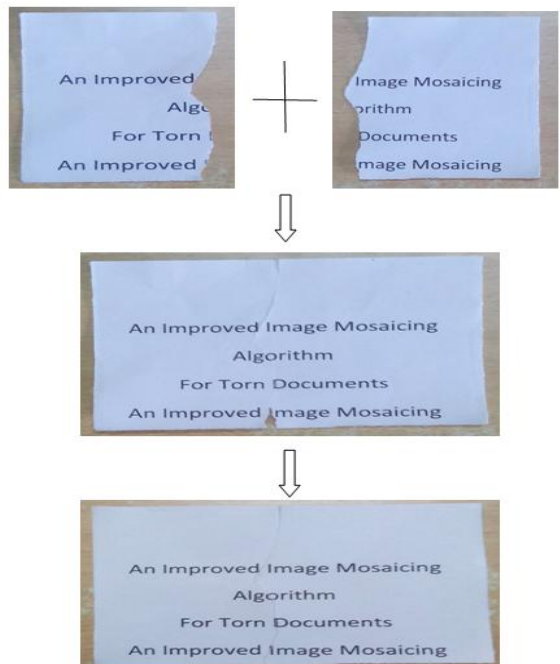

Figure 2: Recovery of Lost Information using character recognition

With the increasing need to recover the information from damaged documents that is visible in various fields, this application proves to be fruitful. Much important information could be recovered with the help of this application. For example, in case of archaeological science which involves study of 100 years, the data is to be documented. Old documents get torn due aging leading to loss of data, which is not affordable. Hence, this application plays a major role in archaeological study.

## II. LITERATURE SURVEY

Mosaicing is the process of stitching two images or fragments to form a single continuous image. Image mosaicing does not consist of a single specific algorithm. For mosaicing of torn documents various combination of algorithm are used for implementing image stitching.

There are various approaches described to implement image mosaicing technique. One of the techniques described where the given document has two torn fragments. The limitation of this technique is that the number of non-uniform side is restricted to one [2]. Also there is a compulsion that the fragments must be placed in such a way that the boundaries of the non-uniform sides always face each other. This limitation is overcome in our project where the non-uniform sides can be more than one and also edge detection is performed so that the non-uniform boundaries can be matched to each other, irrespective of the way they are placed.

In this project the input image is first filtered (i.e. removal of unwanted noise and blurring effect). There is various filtering technique to perform the same but with some limitations. The RANSAC filtering technique is a good filtering technique [1]. But it is a non-deterministic technique since it produces the result with certain probabilities only. This technique is only applicable to certain inliners (i.e. noise with limited range). This technique is not applicable to the images having extreme values of noise. Similarly in Median filter technique which is used for image enhancement, the drawback of this technique is that along with removal of noise it removes certain important fine details of the image. It also involves huge amount of calculations which makes the filtering task more tedious. Hence, in order to overcome these problems we propose a filtering technique known anisotropic diffusion algorithm (also known as Perona-Malik diffusion algorithm).

There are instances where in the torn fragment has two boundaries i.e. an inner boundary and an outer boundary. During the Mosaicing of these fragments the inner boundary is lost and it easily visible that the outer boundaries of the fragments do not match. This problem is overcome in our project by using a polyline simplification algorithm known as Douglas-Peucker (DP).

It is a frequent error that occurs while matching of fragments i.e. we tend to ignore some minute and important details of the matching fragments. To overcome this problem we propose a method of edge detection. The purpose of an edge detector is to capture important events and minute details of the matching fragments. There are various edge detection algorithms. The Marr Hildreth edge detection technique captures the entire fine details and gives multiple edges as output [6]. But this technique has some major drawbacks. The major drawback is that is creates an edge for an edge that do not exist or so called false edges. It also gives high error around the curves of the torn fragments. To overcome these limitations we propose a method that uses canny edge detection. Canny edge detection is a technique that will filter out all the less relevant information and captures the important and minute details.

A document contains several important information for further study and analysis purposes. Loss of this information will halt various important investigation and further studies. In this project, we implement various algorithms for image mosaicing of two torn fragments that will prevent from the loss of information. Image mosaicing is an effective means of constructing a single continuous image from various sub-images. Our algorithm effectively matches the two torn fragments and gives accurate results.

One of the important applications of this topic is in the field of archival study. During the archival study papers are discovered that are fragile and torn due to aging effect. Study and analysis of this ancient discovered paper is an important part of archaeology. Hence, this project presents a simple and efficient approach to match the torn fragments and recover the original document. Hence by studying different papers we can conclude that document reconstruction involves two primary steps i.e. finding the matching torn pieces and then stitching them. In our project we propose various algorithms that will not only match the non- uniform sides of the torn fragments and stitch them out also will give more accurate result (i.e. the reconstructed documents will be very much similar to the original document).

### III. IMPLEMENTATION

The proposed block diagram for mosaicing of two torn documents consists of seven stages. In first stage, filtering would be applied for removal of noise followed by simplifying the rough edges of the torn fragments. This is followed by detection of edges, matching of edges and finally stitching and blending the two torn fragments.

The implementation is summarized using a diagram as shown in figure 3 below.

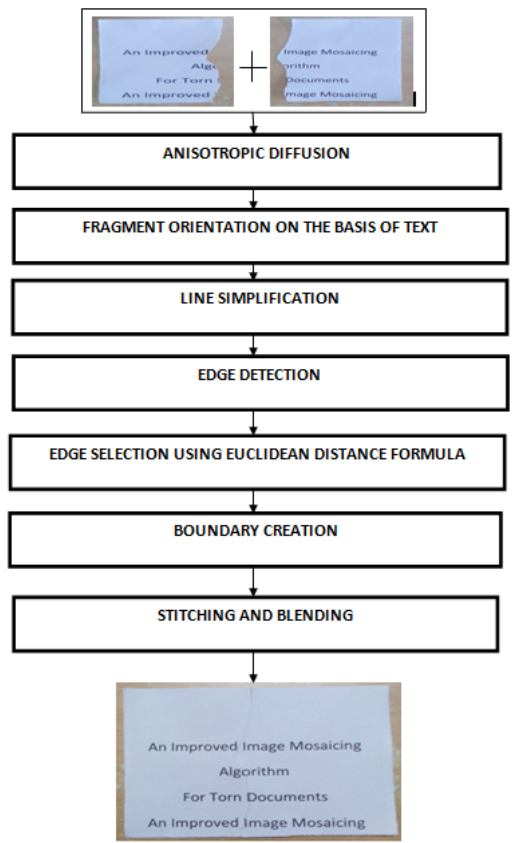

Figure 3: Steps for Recovery from Torn Documents

### IV. IMPLEMENTATION STEPS

#### A. Filtering

This is the first process to be implemented in Mosaicing of two torn fragments. The purpose of this step is to improve the quality of the image. The quality of the image can be improved by removing noise and blurring effect. The algorithm that is being used for filtering process is Anisotropic Filtering algorithm also known as Perona-Malik algorithm. It is a non-linear diffusion algorithm. The main purpose of this step is to eliminate the blurring and localization problem without eliminating any significant part of the image.

#### B. Orientation Of Text

The output image from the previous step is then applied to this step. Often the images are fed randomly to the system, hence their orientation in incorrect. The orientation of these fragments is judge on the basis of the text. The texts of both torn documents are oriented in such a way that they appear similar i.e. either both of them are inverted or both of them are straight. This step helps majorly in detecting the edges that are to be joined together.

#### C. Line Simplification

The torn fragments consist of rough edges also it consists of two boundaries i.e. outer boundary and an inner boundary. Important information may be lost if the outer boundaries are matched instead of inner boundary. To solve this we use line simplification algorithm known as Douglas-Peucker line simplification algorithm. This algorithm reduces the number points in a curve by selecting a tolerance factor (T>0).

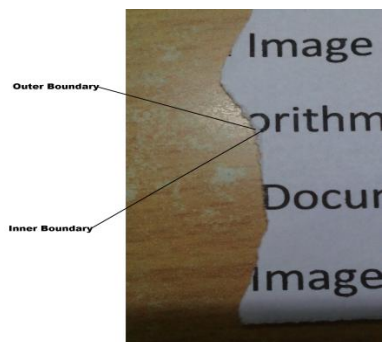

Figure 4: Boundaries associated with an image

Specifically in DP algorithm two endpoints are selected and joined together as an initial rough approximation. Distances are calculated from all vertices of the line segment. If these distances are less than the specified tolerance then the approximation is good else we repeat the steps recursively.

#### D. Edge Detection

The purpose of edge detection is to reduce the amount of irrelevant data in an image, while preserving important information for further processing. Here we are using Canny edge detector for detecting the edges in an image. The Canny edge detection is implemented in the following five steps.

- Smoothening: It is inevitable that all images taken from a camera will contain some amount of noise. Often that noise is mistaken for edges. Hence, noise must be reduced. Therefore the image is first smoothed by applying a Gaussian filter. The kernel of a Gaussian filter with a standard deviation of σ = 1.4.
- Finding Gradients: The Canny algorithm basically finds edges where the grayscale intensity of the image changes the most. These areas are found by determining gradients of the image. Gradients can be calculated by using Pythagoras formula or by Manhattan equation.
- Non-maximum suppression: The purpose of this step is to convert the "blurred" edges in the image of the gradient magnitudes to sharp edges. Basically this is done by preserving all local maxima in the gradient image, and deleting everything else.
- Double thresholding: The edges obtained after non-maximum suppression will probably be true edges but there is a possibility of false edges also. These false edges are created by the noise or color variation. To remove these false edges canny edge detector uses double threshold. Edge pixels stronger than the high threshold are marked as strong; edge pixels weaker than the low threshold are suppressed and edge pixels between the two thresholds are marked as weak.
- Edge tracking by Hysteresis: The strong edges will only be due to true edges in the original image. The weak edges can either be due to true edges or noise/color variations. Thus strong edges can immediately be included in the final edge image. Weak edges are included if and only if they are connected to strong edges.

   *E. Edge Selection*

This is one of the important steps in mosaicing of two torn fragments. For matching of edges of two torn fragments simple Euclidean distance formula is used, where in the distance between two pixels is calculated using the formula as shown in equation (1). This formula is a modification of Pythagoras theorem. In this simple approach, two boundaries of two fragments are considered at a time. Starting from the first pixel on both the boundaries, the distance between them is calculated using the above formula. If the distance between the consecutive pixels comes out to be constant (or in a range nearby a fixed value) then it shows that the boundaries should be stitched together. If the distance between consecutive pixels doesn't satisfy the criteria, then it shows that the boundaries do not match, hence could not be stitched.

$$D = \sqrt{(x2-x1)2 + (y2-y1)2} \qquad (1)$$

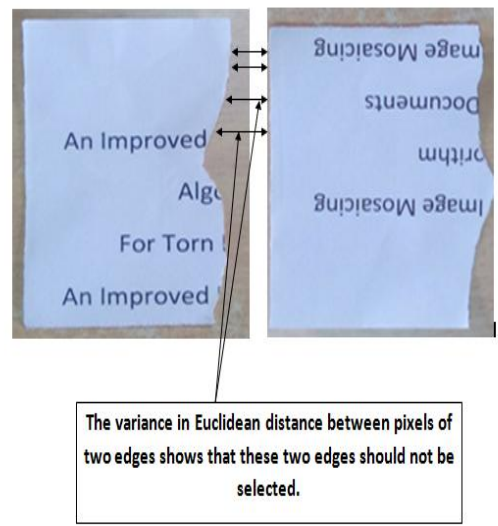

Figure 5: Rejection of two edges on basis of Euclidean distance

In the figure 5, as the distance between consecutive pixels of two edges is not constant, it shows that these edges are not to be selected for merging together

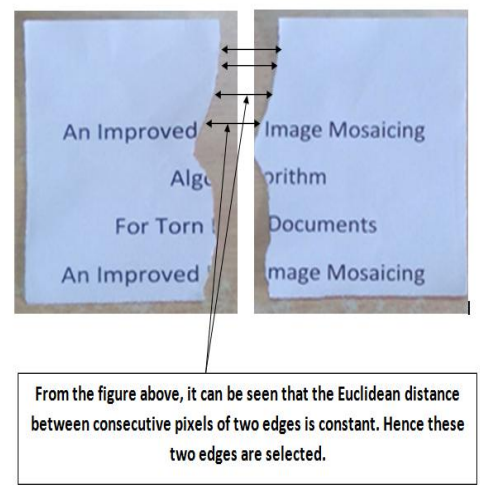

Figure 6: Selecting two edges on basis of Euclidean distance

Whereas, as shown in figure 6, the Euclidean distance calculated between consecutive pixels of two edges is same in all cases. Hence these two edges are chosen for merging together

*F. Boundary-Box Formation*

The boundary box formation plays an important role in stitching of two edges of the two torn fragments. The fragments after proper orientation and after edge selection are subjected to boundary box formation. Where in boxes are created surrounding the fragments at a fixed distance from first pixel.

In this step we, scan the image width row by row and find the first row and last row pixels values where white pixel is encountered mark them as first row and last row.

Likewise, scan the image height column by column and mark them as first column and last column.

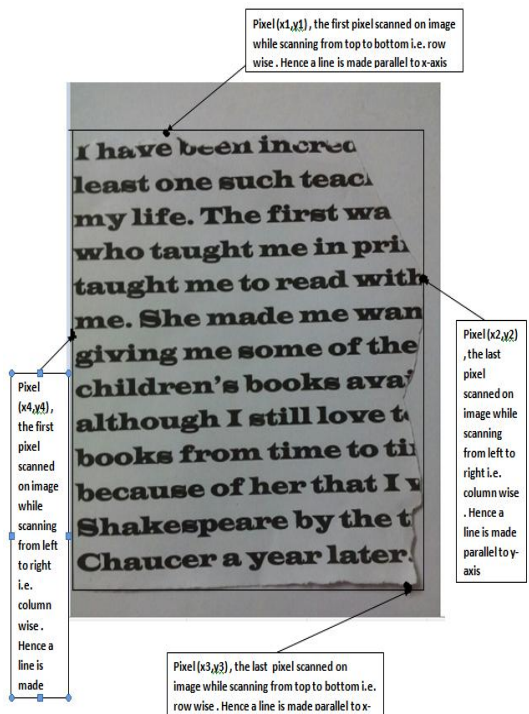

Figure 7: Boundary creation

Draw a rectangular box covering the point of intersection of first row, first column, first row, second column, second row, second column, second row, first column.

*G. Stitching and Blending*

Image stitching or image mosaicing is the most crucial step in recovery of the torn document. After performing all the above steps the matched sides of the torn fragments are stitched together to recover the original document. There are cases where in the recovered original document have distortion and overlapping problems. This problem can be eliminated by using a technique known as Blending. Also character recognition is performed i.e. if a letter is lost in either of the torn fragments; it can be recovered with the help of character recognition technique.

V. RESULT

The algorithm works by taking two fragments of documents. The documents have more than one irregular boundary as shown in the figure below.

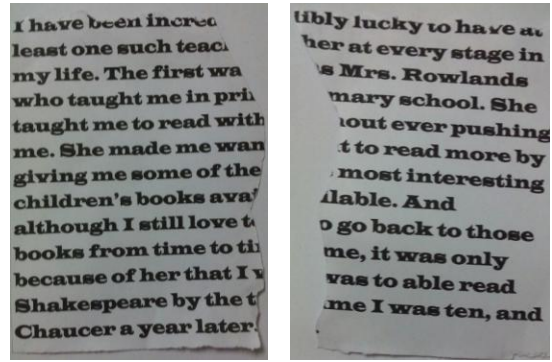

Figure 8: Two Torn pieces of fragments

The algorithm developed works on the above two torn fragments leading to the output as original document as shown below

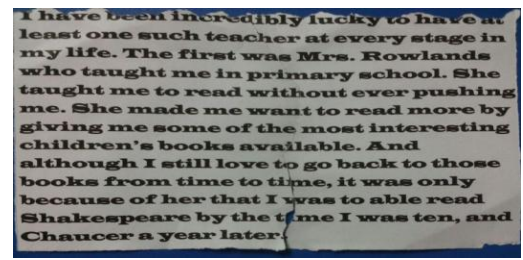

Figure 9: Original document recovered with defect

As it is visible that the information recovered is partial because of missing part of paper leading to incomplete word formation, some information is lost. This loss of information is not acceptable. Hence, in order to overcome this, character recognition is applied, which recognizes the missed characters forming the original word. This gives output as original document.

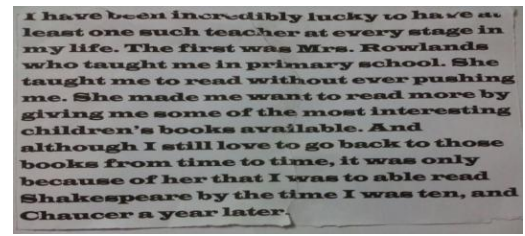

Figure 10: Original Document after Character Recognition.

## VI. CONCLUSION

Manual reconstruction of torn pieces of a document is difficult and a time consuming task. There are various reasons for a document getting torn i.e. by hands or a shredder. In this paper we present a simple yet efficient mechanism for information recovery from torn document. The fragments are scanned and filtering technique is applied for noise removal. The important step is orientation of text followed simplifying the curves. Edge detection and matching of edges technique is proposed for correct stitching of the torn fragments. All the proposed process and algorithm are simple and easy to implement and provides better results.

Further the project can also be extended for joining more than two pieces of torn fragments. Also recognition of more than one character could also be implemented.

## ACKNOWLEDGMENT

We are very grateful to our computer department head Mr. Z.A Usmani. He has been extremely helpful throughout the period of our preparation. We also gratefully acknowledge the support of guides.

.